\documentclass{article}
\usepackage{amsmath,graphicx,mlspconf}
\usepackage{amssymb}
\usepackage{url}
\usepackage{hyperref}
\usepackage{mathtools,tikz,caption}
\DeclareRobustCommand\sampleline[1]{%
  \tikz\draw[#1] (0,0) (0,\the\dimexpr\fontdimen22\textfont2\relax)
  -- (1.2em,\the\dimexpr\fontdimen22\textfont2\relax);%
}
\usepackage{amsthm}

\theoremstyle{definition}
\newtheorem*{definition}{Definition}

\copyrightnotice{979-8-3503-7225-0/24/\$31.00 {\copyright}2024 IEEE}

\toappear{2024 IEEE International Workshop on Machine Learning for Signal Processing, Sept.\ 22--25, 2024, London, UK}
\title{Convexity based pruning of speech representation models}
\name{ Teresa Dorszewski*, Lenka Tětková, Lars Kai Hansen}
\address{Author Affiliation(s)}

\address{%
    Technical University of Denmark \\
   DTU Compute, Section for Cognitive Systems \\
   *{tksc}@dtu.dk
}
\begin{document}
%\ninept

\maketitle

\begin{abstract}
%The abstract should contain about 100 to 150
%words, and should be identical to the abstract text submitted electronically
%along with the paper cover sheet. 
Speech representation models based on the transformer architecture and trained by self-supervised learning have shown great promise for solving tasks such as speech and speaker recognition, keyword spotting, emotion detection, and more. Typically, it is found that larger models lead to better performance. However, the significant computational effort involved in such large transformer systems is a challenge for embedded and real-world applications. Recent work has shown that there is significant redundancy in the transformer models for NLP and massive layer pruning is feasible (Sajjad et al., 2023). Here, we investigate layer pruning in audio models. We base the pruning decision on a convexity criterion. Convexity of classification regions has recently been proposed as an indicator of subsequent fine-tuning performance in a range of application domains, including NLP and audio. In empirical investigations, we find a massive reduction in the computational effort with no loss of performance or even improvements in certain cases.
\end{abstract}
\section{Introduction}
\label{sec:intro}
Self-supervised speech representation models have emerged as highly promising models for many speech-related tasks, such as automatic speech recognition or speech translation \cite{mohamed2022self, superb2021}. There are many different models available with similar performance, differing mainly in pretraining strategies. How these various pretraining strategies influence the representations and how the latent representations develop across the models are still open questions only limited works have explored. In this work, we investigate geometric properties, in particular convexity, of the learned representations and their development across the layers of speech representation models. We also explore how fine-tuning influences the representations and whether knowledge about convex regions in the latent space can help guide the pruning of models to reduce computational complexity and improve downstream performance. 

Recent work has focused on analyzing the word and phoneme content encoded in latent representations using canonical correlation analysis \cite{pasad2021layer, pasad2023comparative}. They show how acoustic, phonetic, and word content develops across layers in several models and relate the word or phoneme content to performance in downstream tasks. They find that using a linear classifier on intermediate frozen layers can outperform classifiers based on the full frozen model, due to more meaningful representations. While our work is inspired by the aforementioned results, it adds a new dimension of analyzing the structure of the latent space directly, by testing geometric properties (in this case convexity). We also fine-tune on intermediate layers directly (i.e. fine-tuning of a pruned model) and show that pruned models can even outperform fine-tuned full models.  

%add short section on pruning of models
Pruning of deep neural networks is a technique that removes redundant neurons not influencing the performance and, hence, reduces the size of a trained network without the need for retraining. There are many types of pruning algorithms \cite{liang2021pruning}. %static or dynamic / element-wise, channel-wise, layer-wise / top-layer dropping (Transformers). 
In this work, we propose a static layer-wise pruning method for transformer models: we prune all layers from a certain point until the end of the network, also called \textit{top-layer dropping} \cite{sajjad2023layers}. Fan et al.\ \cite{Fan2020Reducing} proposed LayerDrop: layer-level dropout during training. They showed that it facilitates robust dropping of layers at test time.
In comparison, our approach allows to use an already existing pretrained model without changing the training procedure. We propose to choose the number of layers for pruning based on the graph convexity score, hence, eliminating the necessity of estimating a suitable number of layers to prune or testing multiple layers.

% Speed-up during inference?
We use a recently developed workflow for measuring the convexity of regions in latent spaces \cite{tetkova2023convex}. Importantly, the authors found evidence that  convexity in pretrained models is related to downstream accuracy of subsequent fine-tuned models. Here we explore this result to establish a pruning strategy. We adapt and apply this metric to speech representation models to improve understanding of these models and the influence of fine-tuning to specific tasks. 

Our main contributions are (i)  in-depth analysis of the latent spaces of state-of-the-art speech representation models, leading to a better understanding of latent representations formed by self-supervised training, (ii) the investigation of changes in the latent representations during fine-tuning to different downstream tasks, and (iii) the development of pruning strategies based on geometric properties of latent representations leading to comparable or better downstream performance with drastically reduced computational complexity.

We find that speech representation models develop distinctive patterns of convex regions, which are similar across most models (except for one). We show that fine-tuning increases the convexity of relevant classes while, at the same time, decreasing the convexity of non-relevant features. Finally, we demonstrate how pruning based on the convexity score can lead to comparable and even better-performing models while decreasing model sizes by up to 70\%, decreasing training times by 20-50\% and inference times by 25-60\%. 
\section{Methods}
\label{sec:methods}
\subsection{Graph Convexity}
We use the graph convexity score introduced by Tětková et al.\ in \cite{tetkova2023convex}. It extends the common definition of convexity for Euclidean spaces to curved manifolds and utilizes sampled data, making it a suitable tool for analyzing high-dimensional latent spaces in a computationally feasible way. Graph convexity is defined as follows: 
\begin{definition}[Graph Convexity]
Let $(V, E)$ be a graph and $A\subseteq V$. $A$ is convex if for all pairs $x, y\in A$, there exists a shortest path $P=(x=v_0, v_1, v_2, \dots, v_{n-1}, y=v_n)$ and $\forall i\in \{0, \dots, n\}: v_i\in A$.
\end{definition}
Converting the problem to data-driven estimators leads to graphs with vertices being the data points and edges defined as Euclidean distances between pairs of data. In curved manifolds, the shortest curve between two points is a geodesic (corresponding to a segment in Euclidean spaces) and that converts to the shortest path in the graph.
Hence, in a simplified way, the graph convexity score is defined as a proportion of the ``well-classified" vertices on the shortest paths between each two points from the same class.
In contrast to \cite{tetkova2023convex}, we calculate the convexity score of the labeled classes and not the decision regions (i.e., we take the target labels as class labels instead of the predicted labels).

We describe the construction in detail: first, we extract representations for all data points after each layer. For each layer separately, we compute Euclidean distances between all representations of data points. We keep only $10$ nearest neighbors and use this as an incidence matrix for a weighted graph. For a pair of points belonging to the same class, we find the shortest path in the graph and compute the proportion of vertices on the path (without the endpoints) that belong to the same class. We average these scores over all pairs of points over all classes. Therefore, the graph convexity score is a number between $0$ and $1$ (where higher is better) computed separately for each model and each layer. Disconnected graphs have a score of 0, while directly connected points (i.e. closest neighbors) have a score of 1. 

We will refer to the graph convexity score as \textit{convexity} for the rest of the paper. The baseline convexity score corresponds to $1/c$, with $c$ being the number of balanced classes.

\subsection{Data}
The dataset used for all experiments is the \textit{speech commands v0.02} \cite{speechcommandsv2}. All audio files are sampled to 16000Hz and zero-padded to a length of 16000 if necessary. From this dataset, we create three datasets to measure different concepts: a word dataset, a speaker dataset, and a phoneme dataset.

\textbf{Word dataset.} The word dataset is the \textit{speech commands} dataset, where we use the \textit{label} information to define the 35 classes (the \textit{\_silence\_} class is excluded). We use the validation set (9982 files) for the convexity analysis and the test set (4890 files) for testing fine-tuned models.

\textbf{Speaker dataset.} For the speaker dataset, we filter the train set from the \textit{speech commands} dataset by the \textit{speaker\_id} to only include speakers that appear more than 50 times. We then split the dataset into a train and test dataset, resulting in 39126 training and 4348 test files with 388 classes. We use the test set for the convexity analysis and testing of fine-tuned models.

\textbf{Phoneme dataset.} For the convexity of phonemes, we create a phoneme dataset based on the validation set of the \textit{speech commands} dataset. Phonemes are defined based on phonetic transcriptions (using X-SAMPA \cite{wells1995xsampa}) of the spoken word, timings for each phoneme utterance are provided by the MAUS web service \cite{kisler2017webmaus} and the corresponding audio parts are extracted. The extracted dataset consists of 33317 audio files covering 35 phonemes. For the convexity analysis, we only include phonemes with more than 1000 occurrences, resulting in 17 phonemes and 25463 audio files. 

\subsection{Models}
\label{sec:models}
We investigate transformer-based models and consider models with different sizes and pretraining strategies to explore the effect of these factors on the representations. We also compare the representations of pretrained and fine-tuned models.

We choose four models based on acceptance in the field and performance on the SUPERB benchmark \cite{superb2021}. In particular, wav2vec2 \cite{baevski2020wav2vec}, wavLM \cite{chen2022wavlm}, HuBERT \cite{hsu2021hubert} and ccc-wav2vec2 \cite{lodagala2023cccwav2vec}. We deploy two sizes of each model: 'base' with 12 layers and embedding size 768, and 'large' with 24 layers and embedding size 1024 (except for ccc-wav2vec2, where no large model is available). Pretrained models are retrieved from \url{huggingface.co}. 

\textbf{Fine-tuning} of the models is done with the Transformers \cite{wolf2020transformers} Trainer module using the training set of \textit{speech commands v0.02} for word classification and the created training set for speaker identification (described above). For fine-tuning, a feature projector and a linear classification layer are added to the pretrained models. Models are fine-tuned to perform audio classification on the 35 available words and speaker identification of 388 speakers. Model performance (accuracy) is reported on the corresponding test set, any hyperparameter tuning and layer selection is based on the validation set. Fine-tuning is done for the full models (i.e., all layers) and pruned models (i.e., only up to a certain layer), based on the experiments conducted (see \autoref{sec:experiments}). Hyperparameters are identical for fine-tuning of all comparable models to minimize co-variates for the analysis. All models are trained with a batch size of 32 for a maximum of 18000 steps with early stopping (based on validation accuracy) and a cross-entropy loss. The learning rate is $2\times10^{-5}$ for full models and $5\times10^{-5}$ for pruned models for the word classification, and $4\times10^{-5}$ for full models and $7\times10^{-5}$ for pruned models for speaker identification. All models were trained successfully with these parameters and converged within the maximum steps. 

\subsection{Experiments}
\label{sec:experiments}
We perform three different sets of experiments. First, we investigate to what extent and where pretrained models form convex regions of relevant information. In the second set of experiments, we explore how fine-tuning to a certain downstream task changes the representations. Lastly, we use insights from the first two experiments to develop an informed pruning strategy and improve fine-tuning efficiency. 

\textbf{Convex Regions in Pretrained Models.} In the first experiment, we extract the latent representations of the three datasets (words, speakers, and phonemes) for all pretrained models after each transformer layer. We measure the convexity and compare how it develops across the models. Based on these results, we choose layers for pruning (see the last experiment).

\textbf{Change of Convexity during Fine-tuning.} Models are fine-tuned to two different downstream tasks: word classification and speaker identification (details are described in \autoref{sec:models}). After fine-tuning, we extract the latent representations of the word, phoneme, and speaker dataset and measure the convexity. We analyze how the convexity changes based on the task and compare convexity scores to downstream accuracies. 

\textbf{Model Pruning Based on Convexity.} For model pruning, we choose the layer with the \textit{best} latent representations in the pretrained model, the best layer is chosen by either the highest convexity score or if the score is not significantly increasing anymore for later layers. We then prune (i.e. delete) all following layers and fine-tune only the remaining layers. For each task, the new final layer is chosen based on the convexity scores of the relevant data (words for the word classification, speaker for the speaker identification), so we train different sizes of models for the different tasks. We compare the model size, downstream accuracy, training time and inference time. Training time is measured only for one training run, but under the same conditions. Inference time is measured 300 times for a random 1s audio input on the GPU. For comparison, we also prune the base models after each second layer to see how pruning at other layers influences the performance.

\section{Results and Discussion}
\begin{figure*}[ht]
    \hspace*{-0.5cm}
    \centering
    \includegraphics[width=\textwidth]{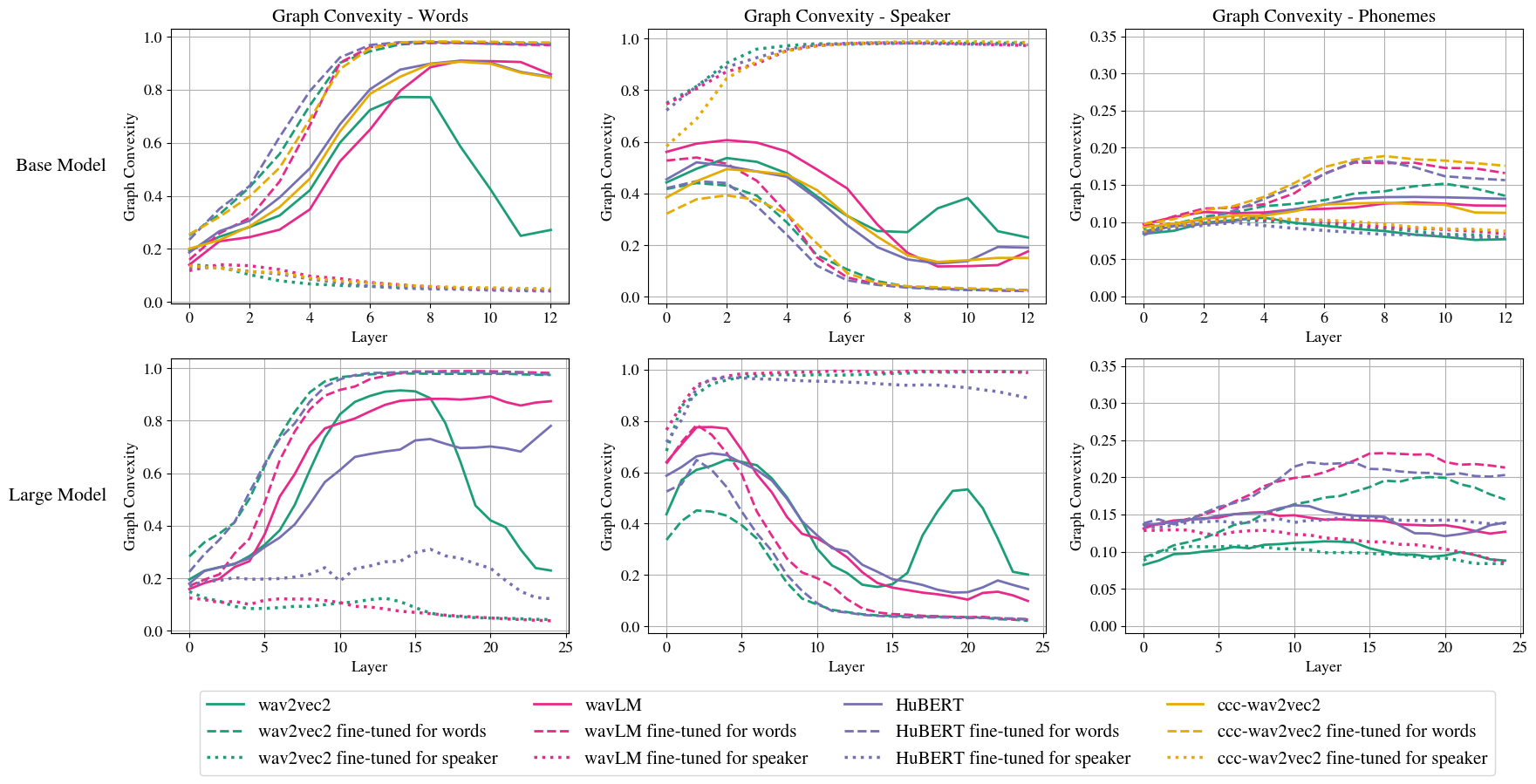}
    \caption{Convexity of latent representations of words, phonemes, and speakers. Evaluated for pretrained (\sampleline{}) and fine-tuned models (word classification: \sampleline{dashed}, speaker identification: \sampleline{dotted}), for base models (upper row) and large models (lower row). Models fine-tuned for word classification show increased convexity for word and phoneme representations and decreased convexity for speaker representations, while models fine-tuned for speaker identification show increased convexity for speaker representations and reduced convexity for word and phoneme representations.}

    \label{fig:convexity_ft}
\end{figure*}
\subsection{Convex Regions in Pretrained Models} 
\autoref{fig:convexity_ft} shows convexity scores for all models, pretrained models are depicted as full lines. The convexity of words increases across the layers (for all models except wav2vec2) while the convexity of speakers generally decreases. The convexity for phonemes stays relatively stable across all layers and is lower than the convexity of words and speakers. The wav2vec2 model has a distinctively different development of the convexity scores compared to all other models. The convexity of words follows an autoencoder-like development, where the convexity increases until layer 8 and then decreases back to levels of early layers. It also has a second peak in speaker convexity in later layers, that other models do not express. 

Very similar behavior for word-content encoding was also observed by Pasad et al.\ \cite{pasad2021layer, pasad2023comparative, pasad2024self}, where especially the distinct auto-encoder-like behavior for words was also observed for wav2vec2 but not for other models. They argue this behavior is caused by different pre-training objectives; in wav2vec2 the objective is to predict acoustic tokens, while the others predict more abstract semantic cluster IDs.

\subsection{Change of Convexity during Fine-tuning} 

We fine-tuned models for word classification and speaker identification, all results can be seen in the exact accuracies for all models can be seen in \autoref{tab:acc_words}. The fine-tuned models for word classification all achieve test accuracy higher than $97\%$ and very similar convexity scores in the later layers. The fine-tuned models for speaker identification all achieve test accuracies of more than $93\%$, except for HuBERT large which only has 69.64\%. This is also reflected in the convexity scores which are lower for HuBERT large (see \autoref{fig:convexity_ft}). We generally observe that the higher the convexity score in the last layer is, the higher the performance. A similar relation between accuracy and convexity was also observed by Tětková et al. in \cite{tetkova2023convex}. 

The fine-tuned models for word classification and speaker identification show very distinct behaviors in their representation space, as shown in \autoref{fig:convexity_ft}. As one would expect, the convexity for the relevant classes increases significantly and is close to 1 for the later layers in all models that have an accuracy higher than $95\%$. The convexity for the non-related class decreases drastically. The convexity of phonemes increases for the word classification model and decreases for speaker identification, which aligns with the expected development as phoneme information can be used for word classification but is not as useful for speaker identification. 

We also observe that the relevant convexity scores converge to a high rate relatively early in the models, for word classification at about half the model and for speaker identification already after 4-5 layers. Based on these observations and the insights from the pretrained models, we hypothesize that later layers in the network may be redundant and not needed for the actual downstream task. We, therefore, conduct the last experiment of pruning the network and only fine-tuning the crucial layers. 

\subsection{Model Pruning Based on Convexity} 
\begin{table*}[ht]
    \centering
    \small
    \begin{tabular}{c||ccccc||ccccc}
     & &  \textbf{word} & \textbf{class.} & &  & & \textbf{speaker} & \textbf{ident.} & & \\\hline \hline
           & Full & Pruned & Acc & Train time & Inference & Full & Pruned & Acc & Train time & Inference\\
    Model  &  Model & Model  & diff. (\%) & diff. (\%)& diff. (\%)&  Model & Model  & diff. (\%)& diff. (\%)& diff. (\%)\\\hline \hline
    wav2vec2 & 98.32 & 98.43 & + 0.11 & - 15.88 & - 22. 73 & 95.17 & 97.79 & + 2.75 & - 38.33 & - 53.25\\
    large & 97.21 & 96.92 & - 0.20 & - 37.48 & - 35.42 & 98.52 & 98.68 & + 0.16 & - 59.08 & - 66.64\\ \hline  
    wavLM & 97.22 & 96.96 & - 0.25 & - 17.33 & - 22.32 & 93.18 & 96.27 & + 3.38 & - 45.30 & - 55.10 \\
    large & 98.86 & 98.66 & - 0.20  & - 25.64 & - 30.36 & 97.14 & 97.53 & + 0.40 & - 66.93& - 64.73\\ \hline
    HuBERT & 97.43 & 96.65 & - 0.80  & - 5.08 & - 18.64 & 94.36 & 92.43 & - 1.80 & - 44.98 & - 52.99 \\
    large & 98.39 & 98.83 & + 0.45  & -17.49 & -30.35 & 69.64 & 96.36 & + 38.37* & - 58.89 & - 65.94 \\ \hline 
    ccc-wav2vec2 & 97.61 & 96.65 & - 0.98 & - 45.09 & -21.81 & 96.20 & 98.55 & + 2.43 & - 21.38 & -52.64 \\ \hline \hline
    Average &  &  & -0.25 & \textbf{-23.42} & \textbf{-25.9} & &  & \textbf{+1.22} & \textbf{-47.84} & \textbf{-58.75} \\
    \end{tabular}
    \caption{Accuracies (in \%, $\sigma$=0.15-0.32\%) and differences in train time and inference time ($\sigma$=0.9-1.1\%) for word classification and speaker identification for full and pruned fine-tuned models. For words classification, the base models are pruned to 8 layers and the large models are pruned to 15 layers. For speaker identification, the base models are pruned to 2 layers and the large models are pruned to 4 layers. On average, the accuracy for word classification decreased by only 0.25\% ($\approxeq \sigma$), while the accuracy for speaker identification increased by 1.22\% (*~excluded). The training time and inference time was significantly reduced in both cases.}
    \label{tab:acc_words}
\end{table*}

We perform static layer-wise pruning on the pretrained models based on the convexity scores for relevant data. We simply delete all transformer layers after the layer with the best convexity score or after which layer the score is not significantly improving anymore. Interestingly, it turns out to be the same layer for all models. For word classification, all base models (12 layers) are pruned to 8 layers and large models (24 layers) to 15 layers. This corresponds to a parameter reduction of 22.48\% and 31.91\% respectively. For speaker identification, the base models are pruned to 2 layers and the large models are pruned to 4 layers. This corresponds to a parameter reduction of 67.38\% and 75.78\% respectively. The pruned models are then fine-tuned for the downstream tasks. 

For the word classification task, the test set accuracies of the pruned model are very comparable to the full model, the exact accuracies are shown in \autoref{tab:acc_words}. Only a slight decrease of 0.25\% was observed on average, while the training time could be reduced by on average 20\% and inference time by 25\%. For the speaker identification task, the pruned models even outperformed the full models on average by $1.2\%$. The training time was reduced to almost half (by 48\% on average) and inference time was reduced by almost 60\%. This clearly shows, how a better understanding of the latent space representations can help with making informed decisions on which layers are useful for certain tasks and lead not only to smaller and more efficient models but even to better performance.
\begin{figure}[ht]
    \centering
    \includegraphics[width=0.49\textwidth]{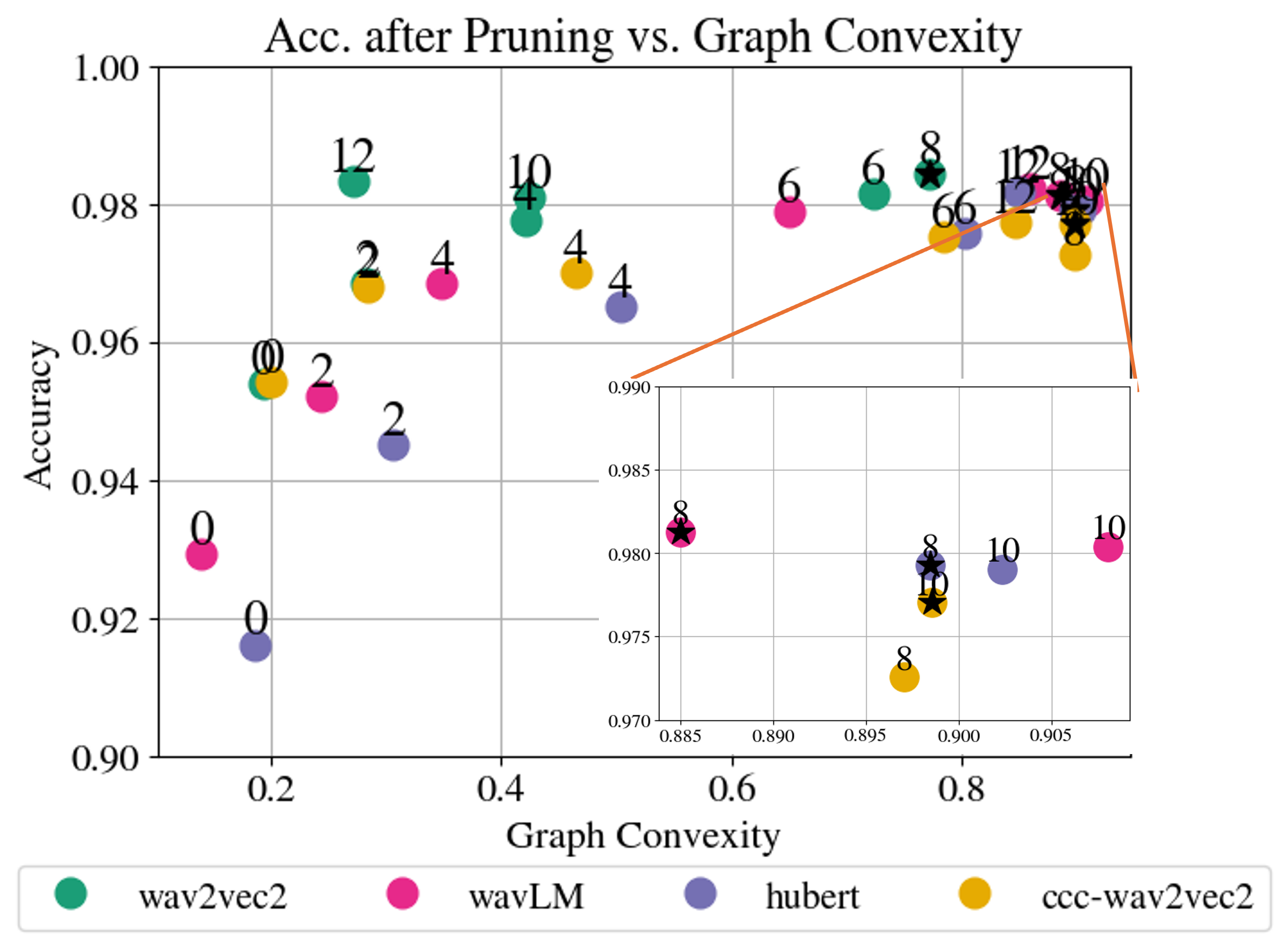}
    \caption{Accuracies for word classification of the pruned base models (number of layers denoted for each point) vs. the convexity score for words for that layer in the pre-trained model. The best performing pruned model is marked with $\bigstar$, which is layer 8 for all models except ccc-wav2vec2.}
    \label{fig:pruning}
\end{figure}

We also prune the base models at every second layer and train it for the word classification task to investigate if the best layer based on the convexity score is actually the best layer to prune the model. The results in \autoref{fig:pruning} show a clear relation between the convexity score of a certain layer and accuracy after pruning to that layer. For three models the best performing pruned model is in fact the one pruned to 8 layers, where the convexity score is also the highest or not significantly increasing for later layers anymore. Eventhough layer 10 is the highest convexity layer for wavLM, HuBERT and ccc-wav2vec2, only for ccc-wav2vec2 the model pruned to 10 layers performs slightly better than the one pruned to 8 layers. This shows, that using the layer after which the convexity doesn't increase substantially anymore instead of using the actually highest convexity layer can be a good strategy as the smaller models have a better/comparable performance. The models pruned to less layers show a slight decrease in performance, however, the accuracy is still 96-98\% for pruning at layers 4, 6 and is still above 92\% for pruning at layer 2 and 0. Even removing all transformer layers and only using the feature extractor still leads to a performance of 70-80\% for all models. This also shows that most of the model can easily be pruned without much of a performance decrease. We omit this experiment for other model and task combinations to save computational resources. Yet, the results for speaker identification in \autoref{tab:acc_words} support the conclusion that pruning at layers with higher convexity scores can actually improve performance, as the accuracy for speaker identification for all except one model significantly increased when using a high convexity layer instead of the full model. 

% \begin{table}[ht]
%     \centering
%     \small
%     \begin{tabular}{c|c||c|c}
%         \# Layers & Acc & \# Layers & Acc \\ \hline \hline   
%          -1 & 79.82 & 6 & 97.58 \\ \hline
%          0 & 94.38 & 8 & \textbf{97.95} \\ \hline
%          2 & 96.20 & 10 & 97.66 \\ \hline
%          4 & 97.09 & 12 & 97.73
%     \end{tabular}
%     \caption{Accuracies (in \%, $\sigma$= 0.20-0.32 (-1 excluded)) for word classification for the model wav2vec2 pruned to different numbers of transformer layers. Pruned to 8 layers performs best. -1 means all transformer layers are dropped.}
%     \label{tab:pruning}
% \end{table} 

Pasad et al.\ \cite{pasad2023comparative} also observed a correlation between the word and phoneme content of the representations of each layer and performance on related tasks. In contrast to this work, they, however, only trained a linear classifier on each layer instead of actually pruning and fine-tuning models. Several other works \cite{liu2021tera, alickovic2023predicting, peng2023structured, zhang2021usefulness} have also found that earlier layers in models have more meaningful representations and that later layers are redundant. Some more complicated pruning strategies \cite{peng2023structured} have also been applied to speech representation models (in particular wav2vec2) with comparable results in size reduction without accuracy decrease. However, these pruning strategies require more optimization and computation. Similar patterns in pruning were observed in NLP task, where performance was maintained within 1\% while pruning up to 9 (of 12) layers and earlier layers were found to be the most critical ones \cite{sajjad2023layers}. 

% "We see that dropping top half of the layers of the network, reduced the number of parameters by 40%, speeding up fine-tuning and inference by 50% with average performance loss between 0.89–2.91 points" -- but I'm not sure what the "average performance loss" and "points" are, it deals with Transformers for text
% "Dropping 4 layers (which gives a speed-up of 33%), RoBERTa achieved a performance close to dropping no layers." 
% "For example QQP maintained performance within 1% on XLNet when 9 layers were dropped (See Table 4). This corresponds to 60% reduction in the number of parameters and 80% reduction in terms of inference time."
% From conclusion: "We conducted experiments using a variety of pre-trained models and using a diverse set of downstream tasks and showed that one can reduce the model size by up to 40%, while maintaining up to 98% of their original performance on downstream tasks. (...) Moreover, we made several interesting observations such as, i) the lower layers are most critical to maintain downstream task performance, ii) certain downstream tasks require as few as only 3 layers out of 12 layers to maintain within 1% performance threshold, iii) networks trained using different objective functions have different learning patterns e.g. XLNet and RoBERTa learns task-specific information much earlier in the network compared to BERT."

\section{Conclusion}
We investigated how convexity of decision regions emerges in self-supervised speech representation models. We find that word and speaker information is mostly encoded in convex regions in specific layers, located in the middle and the beginning of the network respectively. Fine-tuning to word classification and speaker identification increases convexity of the relevant features drastically while decreasing convexity of irrelevant features. Based on the observations of high convexity layers in pre-trained models and converged convexity in early layers of fine-tuned models, we propose that convexity can be used to design a layer pruning workflow, by simply dropping later layers in the transformer stack that do not add further convexity. In empirical investigations, we find massive reduction in the computational effort without loss of performance and interestingly, we found improvements in certain cases. Overall, we find that most transformer layers can be removed with limited or no loss in performance, highlighting the need for more critical view on large transformer models and encouraging the more thorough investigation of latent spaces of audio models to determine the critical and the redundant parts.

\section{Acknowledgements}
This work was supported by the Novo Nordisk Foundation grant NNF22OC0076907 ”Cognitive spaces - Next generation explainability”, the DIREC Bridge project Deep Learning and Automation of Imaging-Based Quality of Seeds and Grains, Innovation Fund Denmark grant number 9142-00001B and by the Pioneer Centre for AI, DNRF grant number P1.
% References should be produced using the bibtex program from suitable
% BiBTeX files (here: strings, refs, manuals). The IEEEbib.bst bibliography
% style file from IEEE produces unsorted bibliography list.
% -------------------------------------------------------------------------
{\small
\bibliographystyle{IEEEbib}
\bibliography{strings,refs}

\begin{thebibliography}{10}

\bibitem{mohamed2022self}
Abdelrahman Mohamed et~al.,
\newblock ``Self-supervised speech representation learning: A review,''
\newblock {\em IEEE Journal of Selected Topics in Signal Processing}, vol. 16, no. 6, pp. 1179--1210, 2022.

\bibitem{superb2021}
Shu{-}Wen~Yang et~al.,
\newblock ``{SUPERB:} speech processing universal performance benchmark,''
\newblock in {\em Interspeech 2021, 22nd Annual Conference of the International Speech Communication Association}. 2021, pp. 1194--1198, {ISCA}.

\bibitem{pasad2021layer}
Ankita Pasad, Ju-Chieh Chou, and Karen Livescu,
\newblock ``Layer-wise analysis of a self-supervised speech representation model,''
\newblock in {\em 2021 IEEE Automatic Speech Recognition and Understanding Workshop (ASRU)}. IEEE, 2021, pp. 914--921.

\bibitem{pasad2023comparative}
Ankita Pasad, Bowen Shi, and Karen Livescu,
\newblock ``Comparative layer-wise analysis of self-supervised speech models,''
\newblock in {\em ICASSP 2023-2023 IEEE International Conference on Acoustics, Speech and Signal Processing (ICASSP)}. IEEE, 2023, pp. 1--5.

\bibitem{liang2021pruning}
Tailin Liang, John Glossner, Lei Wang, Shaobo Shi, and Xiaotong Zhang,
\newblock ``Pruning and quantization for deep neural network acceleration: A survey,''
\newblock {\em Neurocomputing}, vol. 461, pp. 370--403, 2021.

\bibitem{sajjad2023layers}
Hassan Sajjad, Fahim Dalvi, Nadir Durrani, and Preslav Nakov,
\newblock ``On the effect of dropping layers of pre-trained transformer models,''
\newblock {\em Computer Speech \& Language}, vol. 77, pp. 101429, 2023.

\bibitem{Fan2020Reducing}
Angela Fan, Edouard Grave, and Armand Joulin,
\newblock ``Reducing transformer depth on demand with structured dropout,''
\newblock in {\em 8th International Conference on Learning Representations, {ICLR} 2020, Addis Ababa, Ethiopia, April 26-30, 2020}, 2020.

\bibitem{tetkova2023convex}
Lenka T{\v{e}}tkov{\'a}, Thea Br{\"u}sch, Teresa Scheidt, Fabian Mager, Rasmus Aagaard, Jonathan Foldager, Tommy Alstr{\o}m, and Lars~Kai Hansen,
\newblock ``On convex conceptual regions in deep network representations,''
\newblock {\em arXiv preprint arXiv:2305.17154}, 2023.

\bibitem{speechcommandsv2}
P.~{Warden},
\newblock ``{Speech Commands: A Dataset for Limited-Vocabulary Speech Recognition},''
\newblock {\em ArXiv e-prints}, Apr. 2018.

\bibitem{wells1995xsampa}
John~C Wells,
\newblock ``Computer-coding the ipa: a proposed extension of sampa,''
\newblock {\em Revised draft}, vol. 4, no. 28, pp. 1995, 1995.

\bibitem{kisler2017webmaus}
Thomas Kisler, Uwe Reichel, and Florian Schiel,
\newblock ``Multilingual processing of speech via web services,''
\newblock {\em Computer Speech \& Language}, vol. 45, pp. 326--347, 2017.

\bibitem{baevski2020wav2vec}
Alexei Baevski, Yuhao Zhou, Abdelrahman Mohamed, and Michael Auli,
\newblock ``wav2vec 2.0: A framework for self-supervised learning of speech representations,''
\newblock {\em Advances in neural information processing systems}, vol. 33, pp. 12449--12460, 2020.

\bibitem{chen2022wavlm}
Sanyuan Chen et~al.,
\newblock ``Wavlm: Large-scale self-supervised pre-training for full stack speech processing,''
\newblock {\em IEEE Journal of Selected Topics in Signal Processing}, vol. 16, no. 6, pp. 1505--1518, 2022.

\bibitem{hsu2021hubert}
Wei-Ning Hsu et~al.,
\newblock ``Hubert: Self-supervised speech representation learning by masked prediction of hidden units,''
\newblock {\em IEEE/ACM Transactions on Audio, Speech, and Language Processing}, vol. 29, pp. 3451--3460, 2021.

\bibitem{lodagala2023cccwav2vec}
Vasista~Sai Lodagala, Sreyan Ghosh, and Srinivasan Umesh,
\newblock ``Ccc-wav2vec 2.0: Clustering aided cross contrastive self-supervised learning of speech representations,''
\newblock in {\em 2022 IEEE Spoken Language Technology Workshop (SLT)}. IEEE, 2023, pp. 1--8.

\bibitem{wolf2020transformers}
Thomas Wolf et~al.,
\newblock ``Transformers: State-of-the-art natural language processing,''
\newblock in {\em Proceedings of the 2020 conference on empirical methods in natural language processing: system demonstrations}, 2020, pp. 38--45.

\bibitem{pasad2024self}
Ankita Pasad, Chung-Ming Chien, Shane Settle, and Karen Livescu,
\newblock ``What do self-supervised speech models know about words?,''
\newblock {\em Transactions of the Association for Computational Linguistics}, vol. 12, pp. 372--391, 2024.

\bibitem{liu2021tera}
Andy~T Liu, Shang-Wen Li, and Hung-yi Lee,
\newblock ``Tera: Self-supervised learning of transformer encoder representation for speech,''
\newblock {\em IEEE/ACM Transactions on Audio, Speech, and Language Processing}, vol. 29, pp. 2351--2366, 2021.

\bibitem{alickovic2023predicting}
Emina Alickovic, Tobias Dorszewski, et~al.,
\newblock ``Predicting eeg responses to attended speech via deep neural networks for speech,''
\newblock in {\em 2023 45th Annual International Conference of the IEEE Engineering in Medicine \& Biology Society (EMBC)}. IEEE, 2023, pp. 1--4.

\bibitem{peng2023structured}
Yifan Peng, Kwangyoun Kim, Felix Wu, Prashant Sridhar, and Shinji Watanabe,
\newblock ``Structured pruning of self-supervised pre-trained models for speech recognition and understanding,''
\newblock in {\em ICASSP 2023-2023 IEEE International Conference on Acoustics, Speech and Signal Processing (ICASSP)}. IEEE, 2023, pp. 1--5.

\bibitem{zhang2021usefulness}
Shucong Zhang, Erfan Loweimi, Peter Bell, and Steve Renals,
\newblock ``On the usefulness of self-attention for automatic speech recognition with transformers,''
\newblock in {\em 2021 IEEE Spoken Language Technology Workshop (SLT)}. IEEE, 2021, pp. 89--96.

\end{thebibliography}
}
\end{document}